\documentclass[11pt]{article} 
\usepackage[margin=1in]{geometry}
\usepackage{graphicx}
\usepackage{amsmath}
\usepackage{xcolor}
\usepackage{hyperref}
\usepackage{subcaption}
\usepackage{float}
\usepackage{booktabs}
\usepackage[numbers,sort&compress]{natbib}

\hypersetup{
    hidelinks,
    colorlinks=true,
    breaklinks=true,
    urlcolor=blue,
    citecolor=blue,
    linkcolor=blue,
    pdftitle={A Visual Classification Dataset and Model Evaluation for Historical Manuscript Illustrations - Evron, Bar-Asher Siegal, Fire},
}

\title{A Visual Classification Dataset and Model Evaluation for Historical Manuscript Illustrations}

\author{
\begin{tabular}{c}
Yoav Evron$^{1}$, Michal Bar-Asher Siegal$^{2}$, Michael Fire$^{1}$ \\[0.8em]
$^{1}$Faculty of Computer and Information Science, \\
Ben-Gurion University of the Negev, Be'er Sheva, Israel \\[0.5em]
$^{2}$The Goldstein-Goren Department of Jewish Thought, \\
Ben-Gurion University of the Negev, Be'er Sheva, Israel \\[0.8em]
Corresponding author: \texttt{yoavev@post.bgu.ac.il}
\end{tabular}
}

\date{}

\begin{document}

\maketitle

\begin{abstract}
 Historical manuscript illustrations preserve rich visual evidence of past cultures. They depict people, animals, plants, diagrams, music notations, and decorative forms. Although large digitization projects have made many manuscripts available online, the material itself remains difficult to explore at scale. Extraction systems can find illustrations on manuscript pages, but without meaningful categories, large collections remain hard to search and explore. We address this gap by introducing a manually labeled dataset of 15,000 illustrations from manuscripts dating back hundreds of years across 22 categories, and evaluating modern vision models for image classification on this task. The problem is challenging due to stylistic diversity, degradation, and semantic ambiguity, with many images that fit more than one category. We compare fine-tuned CNN and Transformer-based classifiers, zero-shot CLIP, embedding-based classifiers, and direct vision-language models. Results show that fine-tuned image classifiers perform best overall, with ConvNeXt reaching 88.9\% accuracy and 81.3\% macro-F1. Using CLIP embeddings with XGBoost provides a strong alternative. In contrast, zero-shot CLIP and direct vision-language classification perform substantially worse, highlighting the limits of general-purpose models in this domain. Beyond overall performance, the analysis reveals which categories are visually separable and where errors reflect genuine semantic overlap, suggesting that some limitations arise from the taxonomy itself. 
\end{abstract}

\section{Introduction}
\label{sec:intro}
Illustrations embedded in historical manuscripts constitute a rich and important resource for the study of past cultures.  They offer a rare view of how earlier societies represented people, animals, plants, buildings, rituals, and scientific ideas, preserving details that are often hard to recover from text alone. As a result, they support research across many fields, including history, art history, and the history of science \citep{de1994history,epstein2015skies}. At the same time, large-scale digitization efforts have made millions of manuscript pages publicly available \citep{vaticanIIIF,locDigital,blDigitisedManuscripts}. This creates new opportunities for computational analysis, but also a major bottleneck: the sheer volume of visual material exceeds what can be explored manually. Although recent methods can detect and extract illustrations from manuscript pages \citep{evron2026studying, aouinti2022illumination, minisini2024transfer}, extraction alone is not enough. Once these large collections are available, researchers still need effective ways to organize, navigate, and interpret them at scale.

A common way to access such repositories is through automatically generated captions and natural-language search \citep{evron2026studying}. However, this approach remains limited because generated captions can be incomplete or unreliable, especially for degraded and culturally specific manuscript pages. These limitations are partly due to the training data of current vision-language models, which is largely based on modern web imagery and does not fully reflect this domain \citep{rohrbach2019objecthallucination, li2022blip}. More fundamentally, search alone provides only a single access mechanism. Large information systems typically rely on additional organizational layers, such as categorical groupings, to support browsing, comparison, and discovery, particularly when relevant content is unknown in advance.

In this work, we argue that large-scale manuscript illustration repositories require an additional organizational layer: categorization. Rather than replacing search-based access, we complement it with a classification-based approach that organizes extracted illustrations into meaningful categories, such as animals, human figures, battle scenes, and mechanical drawings. This additional layer supports more systematic exploration and interpretation, enabling new forms of discovery and comparison of these images. Despite recent progress in document analysis and illustration extraction \citep{garz2011layout}, standardized evaluation of semantic classification of manuscript illustrations remains limited. Existing work primarily classifies page elements such as text, images, and diagrams, rather than the semantic content of the illustrations themselves.

To address this gap, we introduce a dataset of 15,000 manually verified illustrations from Vatican Library manuscripts across 22 categories, together with a taxonomy, annotation guidelines, and a reproducible evaluation setup. The task is inherently complex: many images are inherently ambiguous and do not fit neatly into a single category. For example, a decorated initial may include animals or human figures, while a map may combine text and plants. These characteristics make both annotation and evaluation difficult, requiring a taxonomy that is stable yet flexible enough for large-scale analysis. In addition, historical manuscript illustrations are often degraded, further complicating the task \citep{hourihane2017classifying, philips2020historical,ziran2023enhancing}.

\begin{figure}[t!]
\centering
\includegraphics[width=\textwidth]{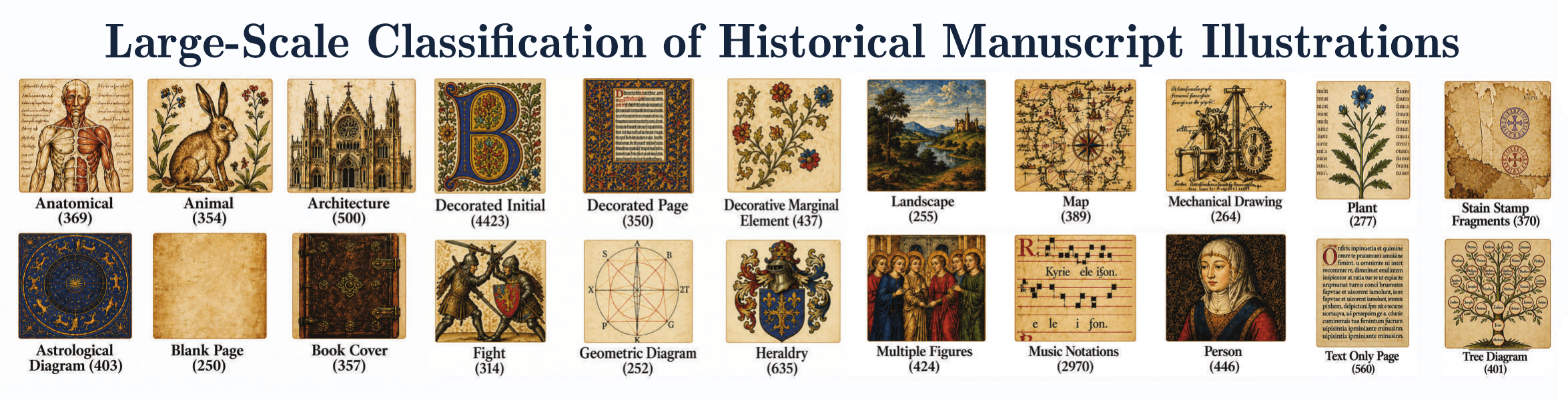}
\caption{A set of 22 illustrations representing the manuscript illustration categories. Each category is shown with the number of images in that category.}
\label{fig:categories}
\end{figure}

\footnote{The representative illustrations were generated by an LLM to respect copyright, and are not samples from the dataset. Original crops are not constrained to square aspect ratios.}

Our proposed taxonomy is designed to provide a practical, reproducible way to organize large-scale collections of manuscript illustrations. It balances historical relevance with usability by assigning each image a dominant category and handling ambiguity through clear annotation rules. We developed the taxonomy through a combination of data-driven analysis, alignment with art-historical systems~\citep{iconclass}, and iterative manual refinement. In addition, we include non-illustration categories, allowing the model to act as an auditing layer and flag likely extraction errors, such as stains, blank pages, or text-only regions. We constructed the dataset using a semi-automatic process that combined generated descriptions with structured human validation.

Using this dataset, we compare major model families, including fine-tuned CNNs, vision transformers, CLIP-based methods, and multimodal approaches. The results show that fine-tuned classifiers perform best overall, with ConvNeXt achieving the strongest performance. In contrast, classifiers built on CLIP embeddings offer a useful trade-off between accuracy and efficiency when embeddings are already available. Our analysis shows that many errors reflect semantic overlap between categories, underscoring the importance of taxonomy design in this domain.

The contributions of this paper are threefold: (1) a reproducible dataset of 15,000 manually verified illustration crops across 22 semantic categories, with annotation guidelines and manuscript-level splits; (2) a unified benchmark comparing supervised classifiers, embedding-based methods, zero-shot approaches, and vision-language models; and (3) a diagnostic analysis of class-level performance, semantic overlap, embedding structure, and taxonomy-related ambiguity.

\section{Related work}
\label{sec:relatedwork}
Large-scale digitization has made historical manuscripts more accessible \citep{vaticanIIIF,ruttenberg2012mass}, opening new opportunities for computational analysis. These collections support research across multiple disciplines. However, they also introduce a major challenge: the visual content in manuscript pages is difficult to locate, organize, and analyze at scale. Most prior work in historical document processing has focused on text-centered tasks such as optical character recognition, handwriting recognition, and page layout analysis \citep{lombardi2020deep, philips2020historical}. More recently, research has begun to address the detection and extraction of visual elements, including illustrations, from manuscript pages \citep{monnier2020docextractor,aouinti2023cv}. However, these efforts primarily focus on detecting where illustrations are located rather than organizing or analyzing their content. 

Closely related work includes S-VED~\cite{buttner2022cordeep} and AnnoPage~\cite{kiss2026annopage}, which jointly localize visual elements and assign broad document-element classes, and docExtractor~\cite{monnier2020docextractor}, which performs page-level segmentation using a general illustration label. These settings either conflate localization and classification or focus on segmentation. Moreover, semantically distinct content such as people, animals, plants, landscapes, and heraldry is often collapsed into a single illustration/image category. These works also do not use manuscript-level splits, which may allow closely related visual styles to leak across train and test sets. They do not report inter-annotator agreement or provide dedicated analyses of class-level confusion, embedding structure, or taxonomic ambiguity. In addition, they do not explicitly audit false-positive extractions, such as stains, text-only regions, or blank pages that an illustration-detection system may mistakenly identify as illustrations.

Image classification is a central task in computer vision, traditionally dominated by convolutional neural networks (CNNs) and, more recently, by Transformer-based architectures \citep{krizhevsky2012imagenet,dosovitskiy2021vit}. Modern CNNs such as EfficientNetV2 \citep{tan2021efficientnetv2} and ConvNeXt \citep{liu2022ConvNeXt} achieve strong performance through improved architectural design and training strategies, while Vision Transformers (ViT) \citep{dosovitskiy2021vit} model images as sequences of patches, capturing broader relationships within the image. Despite their success, it is not clear which model family is best suited for domains such as manuscript illustrations, where images combine fine-grained visual details with higher-level semantic content \citep{khan2023survey}. We therefore treat model selection as an empirical question and compare multiple architecture families.

Vision-language models such as CLIP \citep{radford2021clip} learn a shared embedding space between images and text, enabling zero-shot classification. This is particularly relevant for manuscript illustrations, where many categories are semantic rather than purely visual and may benefit from joint visual-textual representations. Pretrained embeddings can also serve as input to downstream classifiers, such as gradient-boosted trees \citep{chen2016xgboost}. More recent multimodal models, such as LLaVA \citep{liu2023llava}, Qwen2-VL\citep{wang2024qwen2vl}, and InstructBLIP\citep{dai2023instructblip} extend this paradigm by supporting image description and visual question answering. While these models perform well on general image understanding tasks, their effectiveness in specialized historical domains remains unclear, especially when images are stylistically unfamiliar, degraded, or semantically ambiguous. Manuscript illustrations therefore provide a useful case for evaluating whether general-purpose vision-language models are sufficient for this domain.

\section{Dataset}
\label{sec:dataset}
We construct a dataset of 15,000 manually labeled manuscript illustrations across 22 categories. This domain is challenging due to stylistic diversity, historical specificity, and semantic ambiguity. A single image may combine multiple visual functions, such as a decorated initial containing an animal or a diagram embedded within text and ornamentation. Furthermore, we introduce a compact taxonomy, a consistent annotation protocol, and manuscript-level data splits. Together, these components provide a reproducible framework for analyzing how different models behave in this domain.

The dataset is constructed from a large repository of illustrations extracted from digitized manuscripts of the Vatican Library \citep{vaticanIIIF}. The illustrations are obtained using an existing pipeline for illustration extraction \citep{evron2026studying}, which identifies candidate regions containing visual elements within manuscript pages. Our dataset consists of these extracted regions rather than full manuscript pages. This distinction is important, as a single page may contain multiple elements, including text, decorated initials, marginal elements, diagrams, and standalone illustrations. Focusing on cropped regions allows us to isolate the visual content of interest and treat each element as a separate instance. Because the crops originate from high-resolution page scans, their dimensions vary considerably, ranging from small decorated initials to near-full-page illustrations or musical notation. The median resolution is 697$\times$749 pixels (width: 36-4,869; height: 28-4,813), with a median width-to-height ratio of 0.95 (5th-95th percentile: 0.44-2.19).

A central challenge in this task is the lack of a standard taxonomy for manuscript illustration classification. Existing systems, such as IconClass \citep{iconclass}, provide highly detailed iconographic structures, but their level of detail significantly increases data requirements, annotation ambiguity, and evaluation complexity. To address this, we design a compact taxonomy with three goals: it should be \textit{meaningful} for research, \textit{practical} for large-scale learning, and \textit{feasible} to annotate consistently despite ambiguity. The taxonomy is informed by a combination of data-driven exploration, alignment with art-historical concepts, and manual inspection to ensure coverage of recurring visual patterns (see Appendix~\ref{app:taxonomy_construction} for details). It includes 22 categories spanning object types (e.g., \emph{Animal}, \emph{Plant}, \emph{Map}), visual configurations (e.g., \emph{Multiple Figures}, \emph{Fight}, \emph{Decorated Initial}), and non-illustration or artifact categories (e.g., \emph{Text Only Page}, \emph{Stain}), which allow the classifier to support filtering in large-scale illustration extraction pipelines. 

Manuscript illustrations often combine multiple visual and semantic elements within a single image. However, multi-label annotation introduces substantial ambiguity and reduces comparability across models. We therefore define a narrower operational task: assigning each image a single category that best reflects its dominant semantic content, rather than annotating every element present. When multiple elements are present (e.g., a decorated initial containing an animal, or a fight scene containing multiple people), predefined annotation guidelines are used to resolve ambiguity. In general, semantic or figurative content is prioritized when clearly dominant, while manuscript-specific visual forms such as \emph{Decorated Initial} or \emph{Decorative Marginal Element} are selected when they constitute the primary feature of the image. This design reflects a deliberate trade-off between representational richness and evaluation stability, supporting consistent model comparison, reproducibility, and practical collection organization.

To avoid manually annotating from scratch, we adopted a semi-automatic labeling pipeline that combines model-assisted retrieval with human verification. Image captions are generated using LLaVA-1.5-7B (\texttt{llava-hf/llava-1.5-7b-hf}) and indexed with a text-based retrieval system (see Appendix~\ref{app:caption_generation} for the full configuration). We then use category-specific queries to collect candidate examples for each class, enabling efficient identification of both frequent and rare visual types. These candidates are manually verified and labeled in accordance with our annotation protocol. Importantly, automatically generated captions are used only for retrieval and not as ground-truth labels. In practice, captions may omit small objects, miss details such as colors, gestures, and spatial relationships, confuse decorative and semantic elements, or introduce content that is not present in the image \citep{rohrbach2019objecthallucination}. They may also struggle with domain-specific material such as heraldic signs, marginal decorations, and decorated initials. Therefore, we assigned the final labels by manual verification in accordance with the annotation protocol. From the ranked retrieval results, annotators deliberately selected candidates spanning different manuscripts, styles, layouts, and visual configurations, while avoiding near-duplicates and highly repetitive examples.

Annotation followed a written, task-specific protocol refined through iterative analysis of the corpus and involved multiple stages. Initial labeling required substantial manual effort and was performed by a machine-learning researcher specializing in historical illustrations, who developed the taxonomy and classification pipeline and possessed extensive familiarity with both the visual corpus and the computational task. Subsequently, three research assistants reviewed 14,750 images under the direct supervision of a digital humanities expert specializing in historical manuscripts. For each image, reviewers examined its visual content, the proposed label, and the corresponding category definition, flagging 91 cases. Given the substantial effort required for independent annotation of the entire dataset, this stage reviewed proposed labels rather than assigning new labels from scratch. To mitigate potential confirmation bias, the remaining 250 images were independently labeled by the manuscript expert without access to the original annotations, yielding 82.4\% raw agreement and Cohen's $\kappa=0.79$. Thus, every image received a second human verification. Disagreements primarily involved \emph{Decorated Initial} versus \emph{Decorative Marginal Element}, \emph{Architecture} versus \emph{Mechanical Drawing}, \emph{Astrological Diagram} versus \emph{Geometric Diagram}, and \emph{Music Notations} versus \emph{Decorated Initial} (as ornate treble clefs often resemble decorated letters). Finally, a third annotator specializing in computer vision and machine learning adjudicated all flagged cases, resulting in 69 revised labels, which were incorporated into the dataset and all relevant experiments.


\section{Experiments}
\label{sec:experiments}

We compare several modeling approaches for classifying manuscript illustrations, including supervised image classifiers, zero-shot vision-language models, and embedding-based classifiers. All methods use the same class set and the same data split. The dataset is split into training, validation, and test sets using a 70/15/15 split, with manuscript-level grouping applied separately within each category to reduce leakage between splits. To assess robustness to the specific split, we repeat this procedure five times with seeds 42-46, re-drawing the manuscript-level split each time, and report the mean and standard deviation of each test-set metric across the five repeats. Unless stated otherwise, all experiments use a random seed of 42 and a batch size of 16.

\begin{itemize}

\item \textbf{Supervised image classifiers.}
We fine-tune three pretrained torchvision image classifiers: EfficientNetV2-S \citep{tan2021efficientnetv2}, ConvNeXt-Tiny \citep{liu2022ConvNeXt}, and ViT-B/16 \citep{dosovitskiy2021vit}. For each model, the original classification head is replaced with a new linear layer matching the number of manuscript illustration categories. Images are resized to the default input size of the corresponding pretrained model and normalized using its pretrained mean and standard deviation. During training, we use cross-entropy loss with inverse-frequency class weights to handle class imbalance. The models are optimized with AdamW using a learning rate of $1 \times 10^{-4}$ and weight decay of $1 \times 10^{-4}$. A cosine annealing learning-rate scheduler is applied with $T_{\max}$ equal to the maximum number of epochs. Models are trained for up to 12 epochs with early stopping based on validation macro-F1, using a patience of 4 epochs. 
\item \textbf{Zero-shot vision-language models.}
We evaluate CLIP \citep{radford2021clip} in a zero-shot setting using the OpenCLIP implementation with the ViT-B/32 architecture and OpenAI pretrained weights. We test two prompt variants. In \textit{CLIP-short}, each class is represented by its cleaned category name, such as ``animal'', ``map'', or ``plant''. In \textit{CLIP-long}, each class is represented by a longer prompt of the form ``a medieval manuscript illustration of [category]''. For each image, normalized CLIP image embeddings are compared with normalized CLIP text embeddings, and predictions are obtained from the softmax over scaled image-text similarities.
We also evaluate three vision-language models as direct zero-shot classifiers: LLaVA \citep{liu2023llava} (\texttt{llava-hf/llava-1.5-7b-hf}), Qwen2-VL \citep{wang2024qwen2vl} (\texttt{Qwen/Qwen2-VL-7B-Instruct}), and InstructBLIP \citep{dai2023instructblip} (\texttt{Salesforce/instructblip-vicuna-7b}). Each model receives the full category list and generates a single category name using deterministic decoding with a maximum of 20 tokens. LLaVA and Qwen2-VL use an instruction-style prompt, while InstructBLIP uses a question-answer format. Outputs are mapped to category names using normalized text matching.
\item \textbf{Embedding-based classifiers.}
To test whether fixed pretrained representations are sufficient for this task, we train XGBoost classifiers on CLIP-based embeddings. In \textit{CLIP-XGBoost}, we extract normalized CLIP image embeddings using the same ViT-B/32 OpenAI-pretrained CLIP model and train a multi-class XGBoost classifier on these embeddings. In \textit{CLIP-LLaVA-XGBoost}, we first generate an LLaVA caption for each image using a detailed captioning prompt with a maximum of 80 generated tokens. Each caption is then encoded with the CLIP text encoder, and the resulting normalized text embedding is averaged with the corresponding normalized CLIP image embedding. The fused representation is used as input to the same XGBoost classifier.
For both XGBoost-based models, we use a multi-class soft-probability objective with 300 trees, maximum tree depth of 6, learning rate of 0.05, subsampling rate of 0.9, column subsampling rate of 0.9, histogram-based tree construction, and multi-class log-loss as the validation metric. Class imbalance is handled using inverse-frequency sample weights computed from the training labels.
\end{itemize}

All experiments were run under the same software and hardware configuration: Linux, Python 3.10.20, PyTorch 2.5.1, CUDA 12.1, and an NVIDIA GeForce RTX 3090 GPU with 24GB of memory.

\section{Evaluation}
\label{sec:evaluation}
We evaluate how well different model families classify historical manuscript illustrations under the same experimental setup. We compare multiple established approaches, rather than assuming a single modeling paradigm is optimal for this domain. We evaluate model performance using accuracy and macro-averaged precision, recall, and F1 score. Macro-F1 is particularly important in this setting because it gives equal weight to all categories, providing a more reliable view of performance under the strong class imbalance in our dataset, where categories such as music notations and decorated initials are much more frequent than others. Beyond aggregate scores, we analyze performance at the category level by examining per-class results and confusion patterns. This allows us to identify which categories are consistently well separated and which are more confused, providing a more detailed view of model behavior. Finally, we report inference runtime, since the intended application involves large-scale collections with hundreds of thousands of illustrations, where efficiency becomes a key practical consideration. Inference time is measured as average seconds per image on the test set using the same hardware reported above. Unless otherwise stated, timings include the full model-specific inference pipeline, including embedding extraction for embedding-based methods and caption generation for VLM-based methods.

A naive random image-level split is inappropriate for this task, as illustrations from the same manuscript often share visual style and scanning characteristics, leading to overly optimistic performance estimates. We therefore adopt a more appropriate splitting strategy in which, within each category, images from the same manuscript are assigned exclusively to the training, validation, or test set. This prevents style-specific leakage within categories and provides a more realistic evaluation of generalization to unseen manuscripts, reflecting the intended application setting.

Beyond standard evaluation metrics, we analyze the dataset's structure in the CLIP embedding space as a diagnostic tool for understanding class relationships and the role of the taxonomy. The motivation is to examine whether the proposed categories correspond to coherent regions in the learned feature space, and to identify potential sources of confusion that may not be evident from aggregate performance alone. To this end, we project image embeddings into two dimensions using t-SNE \citep{maaten2008tsne} to visualize category-level structure, and compute inter-category similarities using hierarchical clustering \citep{murtagh2012algorithms} to produce a dendrogram. These analyses provide a qualitative view of how categories are organized in the embedding space, and help explain model confusion patterns and analyze errors among visually or semantically similar classes.

\section{Results}
\label{sec:results}
The dataset construction process produced a curated collection of 15,000 manually labeled manuscript illustration images across 22 visual categories including \emph{Anatomical}, \emph{Animal}, \emph{Architecture}, \emph{Astrological Diagram}, \emph{Geometric Diagram}, \emph{Heraldry}, \emph{Landscape}, \emph{Map}, \emph{Mechanical Drawing}, \emph{Music Notations}, \emph{Person}, \emph{Plant}, \emph{Tree Diagram}, \emph{Decorated Initial}, \emph{Decorated Page}, \emph{Decorative Marginal Element}, \emph{Fight}, \emph{Multiple Figures}, \emph{Blank Page}, \emph{Text Only Page}, \emph{Book Cover}, and \emph{Stain Stamp Fragments} (See figure \ref{fig:categories}). Detailed category definitions and annotation guidelines are provided in the annotation protocol.

Table~\ref{tab:results} summarizes the main results.
\begin{table}[t]
\centering
\caption{Performance of the evaluated models on the manuscript illustration classification test set, reported as mean $\pm$ standard deviation over five repeated splits (seeds 42-46). Inference time is the per-repeat mean, in seconds per image.}
\label{tab:results}
\resizebox{\textwidth}{!}{
\begin{tabular}{lccccc}
\toprule
\textbf{Model} & \textbf{Accuracy} & \textbf{Macro Precision} & \textbf{Macro Recall} & \textbf{Macro F1} & \textbf{Inference Time}\\
\midrule
ConvNeXt & \textbf{88.9 $\pm$ 1.5\%}& \textbf{82.2 $\pm$ 1.6\%}& \textbf{81.3 $\pm$ 2.4\%}& \textbf{81.3 $\pm$ 1.8\%}& \textbf{0.005}\\
CLIP-LLaVA-XGBoost& 87.6 $\pm$ 1.1\% & 82.2 $\pm$ 1.0\% & 80.9 $\pm$ 0.8\% & 81.1 $\pm$ 0.8\% &  2.696 \\
EfficientNetV2 & 87.1 $\pm$ 1.6\% & 79.3 $\pm$ 1.6\% & 79.8 $\pm$ 2.1\% & 78.9 $\pm$ 1.6\% & 0.006 \\
CLIP-XGBoost& 85.9 $\pm$ 1.5\% & 79.2 $\pm$ 2.0\% & 76.9 $\pm$ 2.6\% & 77.5 $\pm$ 2.2\% & 0.007 \\
ViT & 85.9 $\pm$ 1.0\% & 76.6 $\pm$ 1.2\% & 75.5 $\pm$ 1.7\% & 75.5 $\pm$ 1.1\% & 0.005 \\
Qwen2-VL Direct & 79.3 $\pm$ 1.1\% & 78.9 $\pm$ 0.4\% & 70.9 $\pm$ 1.2\% & 70.9 $\pm$ 1.0\% & 1.403 \\
CLIP Long Prompt & 58.5 $\pm$ 1.0\% & 53.5 $\pm$ 1.5\% & 59.8 $\pm$ 1.7\% & 53.2 $\pm$ 1.7\% & 0.007 \\
LLaVA Direct & 71.3 $\pm$ 0.7\% & 67.3 $\pm$ 3.2\% & 52.8 $\pm$ 1.5\% & 52.6 $\pm$ 2.0\% & 0.342 \\
CLIP Short Prompt & 58.8 $\pm$ 0.5\% & 54.0 $\pm$ 4.4\% & 32.5 $\pm$ 1.3\% & 33.7 $\pm$ 1.0\% & 0.007 \\
InstructBLIP Direct & 32.9 $\pm$ 1.4\% & 40.5 $\pm$ 3.5\% & 32.6 $\pm$ 1.3\% & 26.4 $\pm$ 1.0\% & 0.277 \\
\bottomrule
\end{tabular}
}
\end{table}

ConvNeXt achieves the best overall performance, reaching 88.9\% accuracy and 81.3\% macro-F1, while also being the fastest end-to-end model. EfficientNetV2 and ViT perform strongly but remain slightly below ConvNeXt across all metrics. Embedding-based classifiers, particularly CLIP-XGBoost and CLIP-LLaVA-XGBoost, achieve competitive results and notably outperform the Vision Transformer baseline, suggesting that high-quality pretrained representations can be highly effective. The CLIP-LLaVA-XGBoost pipeline incurs substantially higher inference time, making it less suitable for large-scale deployment. Off-the-shelf methods generally perform worse than supervised approaches. Among the direct vision-language classifiers, Qwen2-VL performs best, achieving 70.9\% macro-F1, followed by LLaVA at 52.6\% and InstructBLIP at 26.4\%. CLIP zero-shot classification also remains limited under both prompt settings. Within CLIP, prompt design has a clear impact: Longer, more descriptive prompts consistently outperform short category labels by a substantial margin in terms of macro-F1 and macro-recall. In terms of scalability, fine-tuned classifiers are the most efficient, embedding-based methods remain relatively fast, while vision-language approaches are significantly slower and harder to use at scale. Manuscript-specific classes are not uniformly difficult: \emph{Decorated Initial} is visually distinctive and well represented, whereas \emph{Decorative Marginal Element} is more heterogeneous and overlaps with \emph{Decorated Page} and \emph{Decorated Initial}. The strong supervised but weak direct-VLM performance on \emph{Stain Stamp Fragments} further suggests that some manuscript-specific artifact classes are poorly represented in general-purpose multimodal pretraining.

\begin{figure}[t!]
\centering
\includegraphics[width=\textwidth]{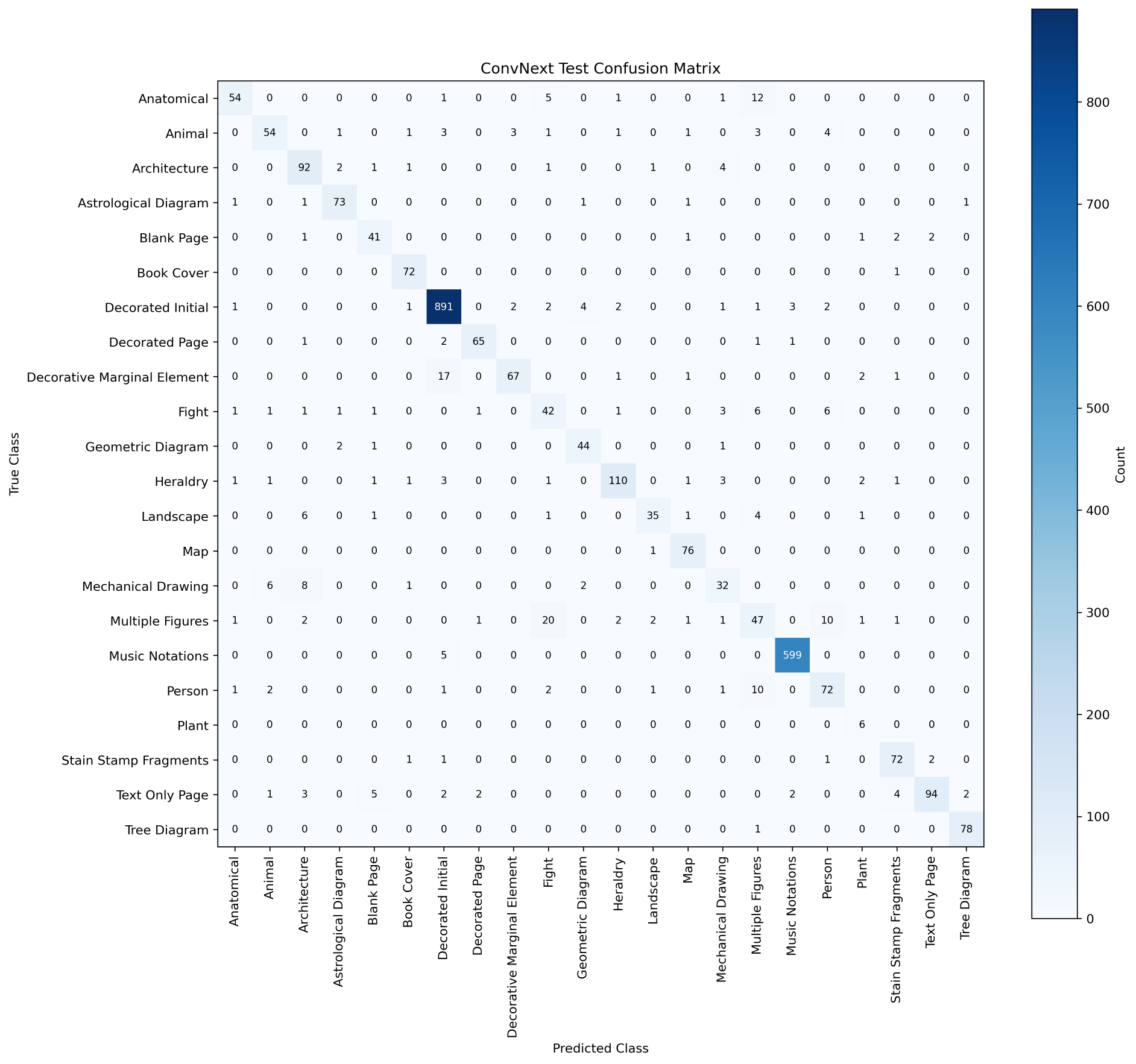}
\caption{Test-set confusion matrix for the best-performing ConvNeXt run (seed 43), selected by macro-F1 for qualitative error analysis.}
\label{fig:confusion_matrix}
\end{figure}

Beyond aggregate performance, the ConvNeXt confusion matrix (Figure~\ref{fig:confusion_matrix}) reveals several recurring ambiguities. \emph{Fight} and \emph{Multiple Figures} are confused in both directions, particularly in crowded scenes where weapons are present but combat is not clearly the dominant action. \emph{Multiple Figures} is also sometimes predicted as \emph{Anatomical} in scenes showing groups around a sick or treated person. Frequent confusion also occurs between \emph{Decorated Initial} and \emph{Decorative Marginal Element}, reflecting their shared ornamental characteristics. In addition, \emph{Architecture} is sometimes predicted as \emph{Landscape} when buildings appear within broader outdoor scenes, or as \emph{Mechanical Drawing} when architectural structures resemble large constructions or machinery. These patterns suggest that several common errors reflect ambiguity in scene interpretation and category boundaries rather than simple visual-recognition failures.

\begin{figure}[t!]
\centering
\includegraphics[width=\textwidth]{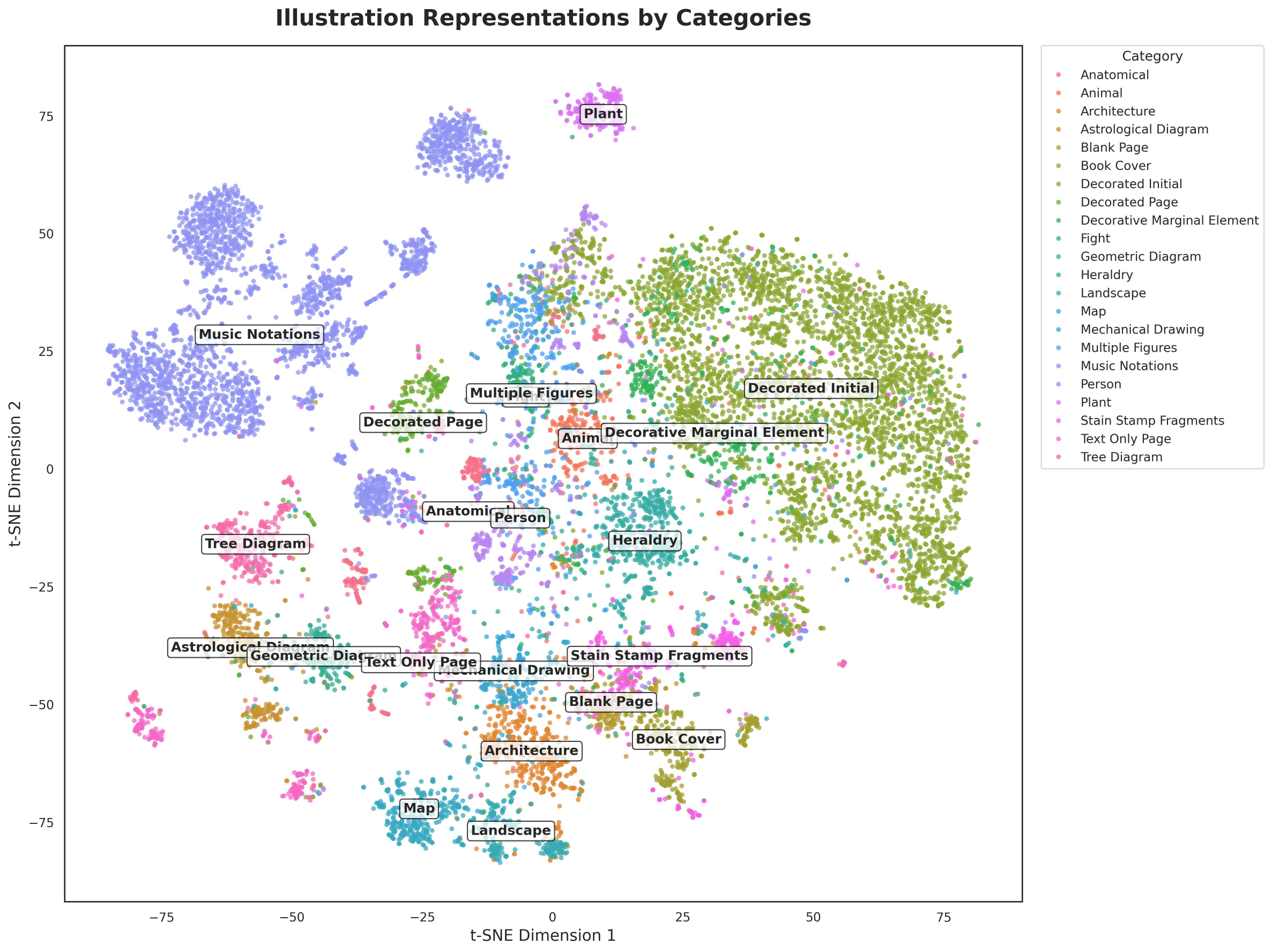}
\caption{t-SNE projection of CLIP image embeddings for the manuscript illustration dataset. Each point represents an illustration and is colored according to its manually assigned category. The visualization provides a qualitative view of category separability and overlap in the visual-semantic embedding space.}

\label{fig:tsne}
\end{figure}

\begin{figure}[t!]
\centering
\includegraphics[width=\textwidth]{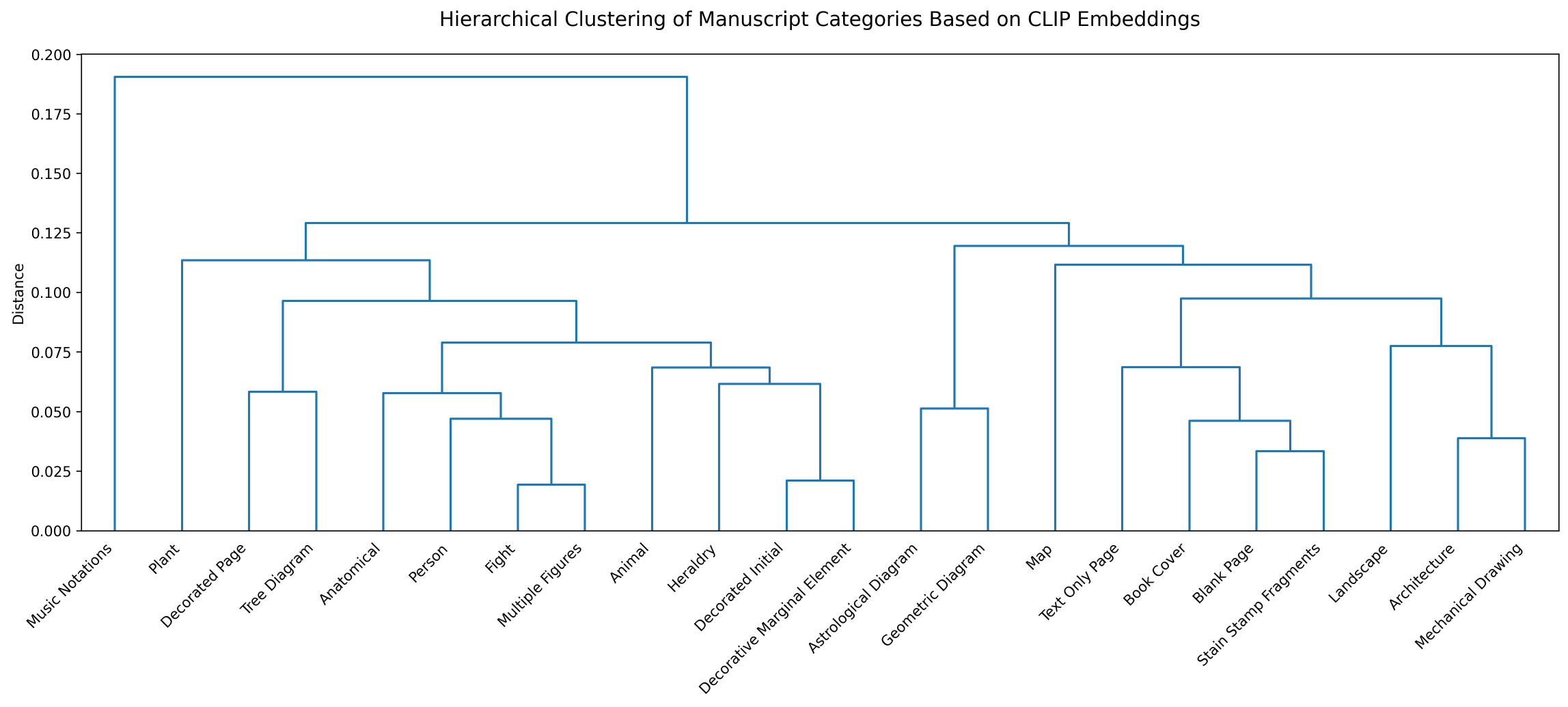}
\caption{Hierarchical clustering dendrogram of the 22 illustration categories based on CLIP embedding correlations. The dendrogram summarizes category proximity in the embedding space and helps interpret visual or semantic relationships between categories.}
\label{fig:category_clip_hierarchical_clustering_dendrogram}
\end{figure}

The embedding analysis supports this interpretation. In the CLIP t-SNE projection (Figure~\ref{fig:tsne}), visually distinctive categories, such as \emph{Music Notations}, \emph{Map}, and \emph{Decorated Initial}, form relatively coherent clusters, while overlapping categories are more mixed. Similarly, hierarchical clustering of category-level correlations (Figure~\ref{fig:category_clip_hierarchical_clustering_dendrogram}) places related categories close to one another, including \emph{Person}, \emph{Multiple Figures}, and \emph{Fight}, as well as diagrammatic classes such as \emph{Astrological Diagram}, \emph{Geometric Diagram}, and \emph{Tree Diagram}. This indicates that, despite its weaker zero-shot classification performance, CLIP embeddings still capture meaningful semantic relationships between manuscript illustration categories. However, the continued weakness of CLIP on manuscript-specific categories, such as \emph{Decorated Marginal Element}, suggests that these errors reflect a deeper domain mismatch rather than prompt sensitivity alone.

\begin{figure}[t!]
\centering
\includegraphics[width=\textwidth]{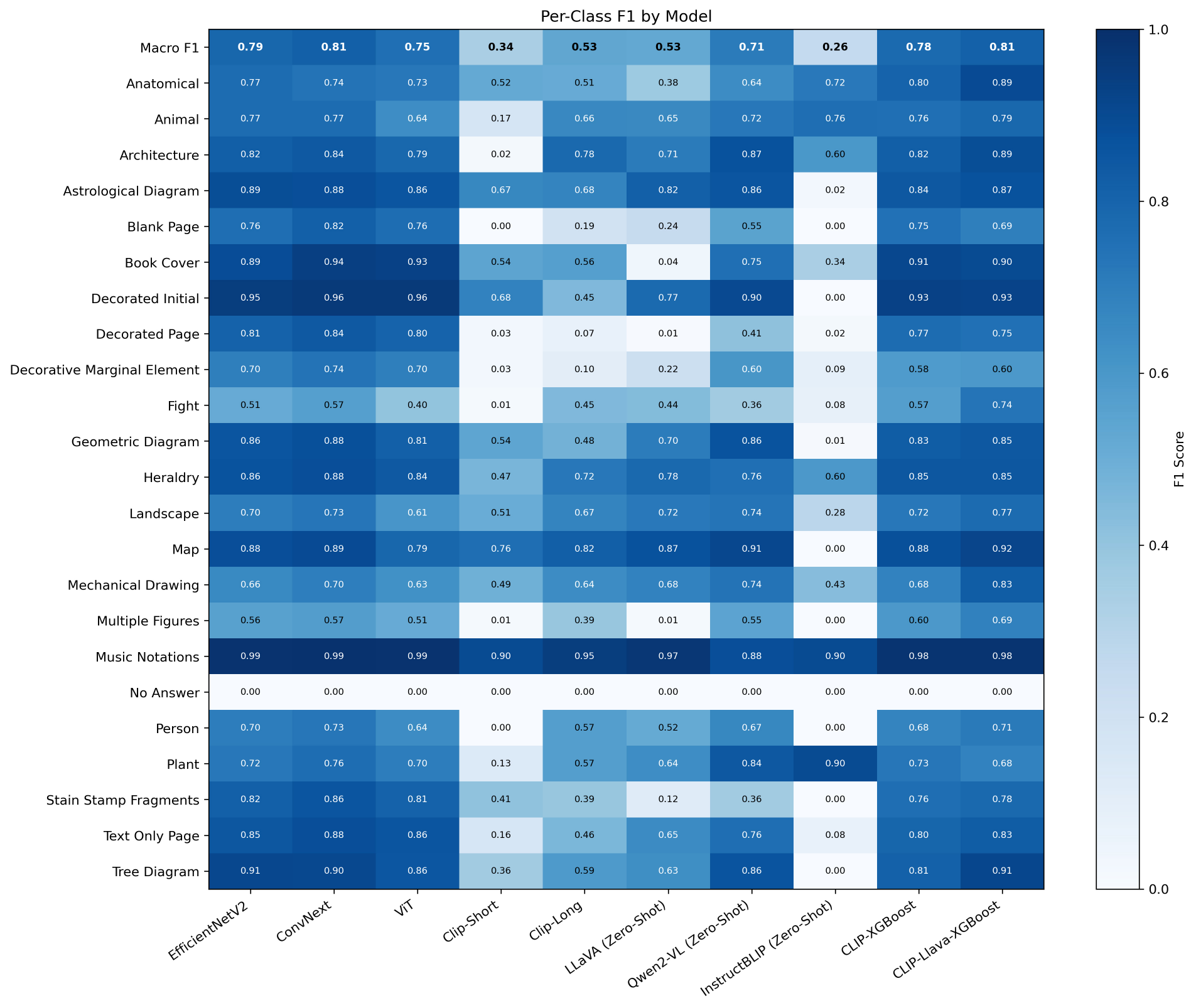}
\caption{Per-class F1-score heatmap for the evaluated models across the 22 manuscript illustration categories. The heatmap highlights categories that are consistently easy or difficult to classify and reveals class-specific differences between model families.}
\label{fig:per_class_f1_heatmap}
\end{figure}

\section{Discussion}
\label{sec:discussion}
This study addresses the classification of extracted illustrations from historical manuscripts, a challenging task due to high visual diversity and overlapping categories. First, the proposed taxonomy provides a practical structure for organizing large collections of manuscript illustrations. Rather than capturing the full iconographic richness of historical material, it defines a reproducible evaluation setting that balances interpretability, coverage, and annotation feasibility. The results show that this structure is effective, while also revealing that some category boundaries are inherently ambiguous. It remains open to extension in future work. A further limitation is that candidate collection relied on model-assisted retrieval, potentially introducing selection bias toward visually or textually salient examples despite the deliberate sampling of diverse manuscripts, styles, layouts, and visual configurations. Furthermore, although the Vatican Library collection spans diverse periods, languages, styles, and visual traditions, it represents a single institutional source, and extending the benchmark to additional collections remains a direction for future work.

Second, the single-label annotation protocol enables consistent and comparable evaluation across models, but at the cost of simplifying the multi-label nature of manuscript imagery. As a result, some classification errors reflect this simplification rather than clear model failures. This suggests that extensions such as multi-label or hierarchical annotations, or decomposing images into multiple objects, could better capture the domain's complexity. In addition, our error and embedding analyses show that many misclassifications occur between semantically overlapping categories. Categories that are close in the representation space are also more likely to be confused. This indicates that performance is closely tied to the structure of the taxonomy, and that some errors reflect meaningful ambiguity in the data rather than incorrect predictions.

Third, the comparison between models shows that task-specific training is highly effective for this task. ConvNeXt achieves the highest mean performance, while embedding-based methods provide a strong alternative and outperform the Vision Transformer baseline. The results also show clear limitations of zero-shot approaches. Among the direct vision-language models, Qwen2-VL performs substantially better than LLaVA and InstructBLIP, yet all remain below supervised classifiers, highlighting the gap between general-purpose models and this domain-specific task. These errors are concentrated in manuscript-specific categories, such as decorative elements, which are underrepresented in general-purpose training data, highlighting a clear mismatch between pretraining data and this domain. The large gap between short and long CLIP prompts further highlights how sensitive zero-shot performance is to prompt design. In addition, inference time is an important practical factor: ConvNeXt is the strongest default choice, while CLIP-based classifiers are useful when embeddings are already available.

Taken together, these results show that manuscript illustration classification is not just a prediction task. Dataset design, annotation choices, and taxonomy structure also shape it. While the study demonstrates that large-scale classification is feasible, many of the main challenges are still conceptual. In particular, questions remain about how to define categories and how to handle ambiguous or multi-content images. This work is therefore a novel contribution and a starting point. Beyond ranking models, the benchmark identifies which categories are visually separable, which benefit most from domain-specific supervision, and which remain challenging because of genuine taxonomy overlap. Future research can expand the dataset, explore multi-label or hierarchical approaches, and evaluate newer models, helping build more practical tools for large-scale manuscript analysis.

\section{Conclusions}
\label{sec:conclusions}
We introduced a dataset and evaluation setting for the classification of illustrations in historical manuscripts, together with a practical taxonomy and an annotation protocol. This work addresses a gap between illustration extraction and large-scale visual organization. We build a dataset of 15,000 images labeled into 22 categories and compare multiple modeling approaches. The results show that fine-tuned image classifiers provide the most reliable performance in this domain, with ConvNeXt achieving the best overall balance of accuracy, macro-F1, and inference speed. Embedding-based approaches remain useful when CLIP representations are already part of the pipeline. However, they are not the most efficient end-to-end solution in our setting. More importantly, the results show that performance is shaped not only by model choice but also by the taxonomy's structure and the evaluation protocol. Many errors arise from semantic overlap between categories and reflect ambiguity in the data rather than clear model failures. Overall, improving performance in this domain requires not only better models but also clearer problem formulation. The dataset and framework introduced in this work provide a basis for further research on large-scale analysis of manuscript illustrations and similar visual domains.

\section{Data and code availability}

The manuscript images used in this study are publicly accessible through the Vatican Library's digital collection but cannot be redistributed directly. To ensure reproducibility, upon publication of this work, we will publicly release all annotations and reconstruction metadata, including manuscript and page identifiers, crop coordinates, category labels, and split assignments. We will also release an easy-to-use reconstruction script that retrieves the original scans through the Vatican Library's official IIIF service and automatically recreates all 15,000 illustration crops. The reconstruction code, model benchmark code, evaluation outputs, annotation protocol, and best-performing trained models will likewise be made publicly available upon publication.

\section*{Acknowledgments} 
We gratefully acknowledge the generous support of The Soref-Breslauer Texas Foundation, whose contribution made this research possible.


\phantomsection


\section{Appendix}
\subsection{Taxonomy Construction}
\label{app:taxonomy_construction}
The taxonomy was constructed through a hybrid process combining data-driven exploration, domain-informed design, and manual validation. Rather than adopting an existing iconographic taxonomy directly, our goal was to define a compact and operational set of categories suitable for supervised classification of extracted manuscript illustrations at scale. Overall, the resulting taxonomy reflects a balance between three complementary factors: statistical structure in the image embedding space, domain-informed interpretability, and empirical coverage of the extracted corpus. It should therefore be understood as a practical research taxonomy rather than a complete ontology of manuscript imagery.

First, we explored the visual structure of the corpus using CLIP image embeddings~\cite{radford2021clip}. Mini-batch K-means clustering was applied to a large sample of extracted illustrations, and the number of clusters was selected using the elbow method. We used $k=40$ as a practical balance between visual granularity and interpretability. To interpret the resulting clusters, we manually inspected random samples from each cluster and identified recurring visual patterns. Second, the candidate categories were compared with established art-historical and iconographic concepts, particularly those represented in systems such as IconClass~\cite{iconclass}. However, we intentionally reduced the granularity of such systems. The purpose of the taxonomy was not to reproduce the full iconographic richness of manuscript studies, but to create a reproducible evaluation structure that could be annotated consistently and used for model training. Third, we performed a coverage validation step by manually inspecting a random sample of 10,000 extracted illustrations. This step was used to verify that common visual types were represented, identify missing or underrepresented classes, and refine the boundaries between visually overlapping categories. It also motivated the inclusion of practical non-iconographic categories, such as blank pages, text-only pages, stains, stamps, and fragments, which are important for real-world extraction pipelines.

\subsection{Caption Generation Configuration}
\label{app:caption_generation}

Captions used during dataset construction were generated with LLaVA-1.5-7B (\texttt{llava-hf/llava-1.5-7b-hf}) using the prompt ``Describe this image in as much detail as possible, mentioning elements, colors, objects, relationships, and any emotions.'' and a maximum of 80 generated tokens.

\subsection{Additional Results}
\label{app:additional_results}

This section reports additional visual analyses that complement the main experimental results. 

\end{document}